\newcommand{\etal}{\textit{et al. }}

\documentclass[10pt,twocolumn,letterpaper]{article}
\usepackage{cvpr}

\usepackage{graphicx}
\usepackage{amsmath}
\usepackage{amssymb}
\usepackage{booktabs}
\usepackage{algorithm}
\usepackage{algorithmic}
\usepackage{multirow}
\usepackage{cases}
\usepackage{booktabs}
\usepackage{color}
\usepackage[pagebackref,breaklinks,colorlinks]{hyperref}

\usepackage[capitalize]{cleveref}
\crefname{section}{Sec.}{Secs.}
\Crefname{section}{Section}{Sections}
\Crefname{table}{Table}{Tables}
\crefname{table}{Tab.}{Tabs.}

\def\cvprPaperID{*****} 
\def\confName{CVPR}
\def\confYear{2023}

\begin{document}

\title{One-stage Modality Distillation for Incomplete Multimodal Learning}  

\author{Shicai Wei \quad\quad\quad  Chunbo Luo \quad\quad\quad  Yang Luo  \\
School of Information and Communication Engineering \\
University of Electronic Science and Technology of China\\
{\tt\small shicaiwei@std.uestc.edu.cn  \{c.luo, luoyang\}@uestc.edu.cn }
}
\maketitle

\begin{abstract}
Learning based on multimodal data has attracted increasing interest recently. While a variety of sensory modalities can be collected for training, not all of them are always available in development scenarios, which raises the challenge to infer with incomplete modality. To address this issue, this paper presents a one-stage modality distillation framework that unifies the privileged knowledge transfer and modality information fusion into a single optimization procedure via multi-task learning. Compared with the conventional modality distillation that performs them independently,  this helps to capture the valuable representation that can assist the final model inference directly. Specifically, we propose the joint adaptation network for the modality transfer task to preserve the privileged information. This addresses the representation heterogeneity caused by input discrepancy via the joint distribution adaptation. Then, we introduce the cross translation network for the modality fusion task to aggregate the restored and available modality features. It leverages the parameters-sharing strategy to capture the cross-modal cues explicitly. Extensive experiments on RGB-D classification and segmentation tasks demonstrate the proposed multimodal inheritance framework can overcome the problem of incomplete modality input in various scenes and achieve state-of-the-art performance.

\end{abstract}

\section{Introduction}

\label{introx}

Recently, multi-modal learning has attracted growing attention from academia and industry and promoted the development of many applications, such as action recognition~\cite{action_multi_2,action_multi_all_1} and semantic segmentation~\cite{rgbd-seg-1,rgbd-seg-2,rgbd-seg-3}. However, most successful multimodal models assume that the inference modalities are always complete. This is difficult to meet in practice due to the limitation of devices~\cite{d2d} or user privacy~\cite{privacy1}. Thus, how to bridge the gap between the ideal training scenarios with complete modalities and the real-world deployment scenarios with incomplete modalities is of great significance for multimodal tasks.



To handle the challenge of incomplete multimodal inference, imputation-based methods and distillation-based methods are proposed to address the challenge of incomplete multimodal inference, respectively. Specifically, the modality data available only at training time is referred to as the privileged modality~\cite{wei2023privileged}. Here, we denote the modality data available at both training and inference as the ordinary modality. The imputation-based methods aim to estimate the privileged modality sample by imputation algorithms and then combine it with the ordinary ones as the input of the pre-trained multimodal model~\cite{MC1,MC2}. Because the sample reconstruction is complicated and unstable, this would harm the primary task at hand\cite{GAN-privilege,gan-issue1,gan-issue2}. Thus, the distillation-based methods are proposed to reduce the estimation problem from the sample space to the feature space~\cite{MH1,mmanet,dmr,wei2023msh}. They first train a hallucination model to learn the privileged representation knowledge by distillation algorithms and then fuse it and the model trained with the ordinary modality data to make the final decision.

Hypothesizing that the distillation-based methods are fruitful, we can conclude that the key of handling incomplete modalities is to restore the valuable privileged information from the ordinary modality. Here, the `valuable' means the information that can advance the performance of the ordinary modality. For example, pixel information is not necessary for inference and thus distillation methods can only focus on the representation knowledge. Besides, while the distillation-based methods alleviate the complexity of reconstruction tasks effectively, they still focus on restoring complete privileged representation from the ordinary modality. Nevertheless, due to the intrinsic modality discrepancy, it is infeasible to conduct accurate reconstruction. Note that slight differences can lead to a huge impact on the performance of the model~\cite{a-attack1,a-attack2}. From this perspective, we should pay attention to the valuable instead of complete representation knowledge.



Motivated by this, we propose a one-stage modality distillation framework (OS-MD) that joints the hallucination model learning and the modality information fusion into the same optimization procedure via multi-task learning. Compared with the conventional modality distillation that performs them independently, joint optimization helps to extract the privileged representation that can assist the final model inference directly. Instead of simply adding the loss functions of hallucination learning and modality fusion, we introduce the joint adaptation network and cross translation network for privileged knowledge transfer and modality information fusion, respectively. The joint adaptation network guides the hallucination model to learn the representation of the privileged model by matching the joint distribution of the sample representation and sample label. On the one hand, this contributes to a simpler optimization objective compared to the feature-based distillation that matches feature values. On the other hand, this preserves the conditional distributions between sample representation and final prediction output that is crucial for the model inference~\cite{wei2025boosting,wei2024gradient,discri3}. The cross translation network explicitly fuses multimodal representation by sharing parameters. Compared with existing distillation-based methods that fuse information at the output layer, this ensures the OS-MD model learns the intermediate cross-modal cues. We summarize the main contributions of this work as follows:

\begin{itemize}
    \item We propose a one-stage modality distillation framework (OS-MD) to joint the hallucination model learning and the multiple modality information fusion into the same optimization procedure via multi-task learning, so that these procedures can negotiate with each other to capture valuable privileged knowledge.
    \item We propose the joint adaptation network to transfer privileged representation knowledge to the hallucination model by matching the marginal and conditional distributions. This helps to reduce the optimization difficulty of hallucination learning while preserving the crucial information of privileged modality for multiple modalities fusion.
    \item We introduce the cross translation network to translate the sample representation of the hallucination model into that of the representation into that of the ordinary model via the shared parameter. This helps to extract the intermediate cross-modal cues that are important for multimodal learning.
\end{itemize}

\section{Related Work}

\subsection{Imputation-based methods}
The imputation-based methods reconstruct the pixel information of the privileged modality from the ordinary modality image for the pre-trained multimodal classification model to handle the missing modality. For example, Jiang~\etal utilize the CycleGAN~\cite{cyclegan} to generate the MRI image from the CT image to aid mediastinal lung tumor segmentation~\cite{pixel-based-3}. Pan~\etal impute the missing PET images based on their corresponding MRI scans using a hybrid generative adversarial network and leverages them to aid the diagnosis of brain disease~\cite{pixel-based-2}. Because the pixel-level generation is complex, it may limit the primary task at the hand, such as classification and segmentation~\cite{gan-issue1,GAN-privilege}.

\subsection{Distillation-based methods}
The Distillation-based methods are proposed to train an extra model to reconstruct the representation information of the privileged modality and fuse it with the model trained with the ordinary modality. Hoffman~\etal first propose to hallucinate model to mimic the privileged representation by the distillation methods and introduce the hints distillation to hallucinate the depth representation for the RGBD object detection~\cite{MH1}. Saurabh~\etal train the hallucinate hallucinated representation of missing spectral data for multi-spectral land cover classification~\cite{miss-hall2}. Crasto~\etal further extend the hallucination architecture for the video action recognition to model the motion flow explicitly to improve its performance~\cite{ADDA}. Li~\etal introduce the dynamic-hierarchical attention distillation to hallucinate the SAR image feature from RGB images to aid the land cover classification~\cite{miss-dhad}. Garcia~\etal propose the adversarial discriminative modality distillation to mimic the representation distribution of the privileged model to assist the multimodal action recognition~\cite{GAN-privilege}. In summary, these works mainly focus on privileged knowledge learning while ignoring the fusion of modality information. Noting that the latter is exactly the cruciality for the task at hand.

\subsection{Multi-task learning}
\label{mul}

Multi-task learning is proposed to improve data efficiency and reduce overfitting by leveraging auxiliary information~\cite{multitask-learning-survey}. The existing multi-task learning methods have often been partitioned into two groups: hard parameter sharing and soft parameter sharing. The hard parameter sharing is the practice of sharing model weights between multiple tasks, so that each weight is trained to jointly minimize multiple loss functions. Although these methods can reduce the overfitting significantly, they may lead to negative transfer and harm the performance of each task when the correlation between multiple tasks is low. Thus,  soft parameter sharing is proposed to provide separate weights for the individual models of each task. Then the distance between model parameters is added to the joint objective function to leverage the complementary information from different tasks to assist model optimization. Besides, the previous study has demonstrated that the balance of multiple losses has a great influence on the performance of multi-task learning~\cite{multitask-learning-survey}. And a large number of methods have been proposed to tackle this problem. For example, Zhao~\etal propose to manipulate gradient norms overtime to control the training dynamic~\cite{grad}. Kendall~\etal propose to modify the loss functions in multi-task learning using task uncertainty~\cite{wei2025improving}. As an alternative to using task losses to determine task difficulties, Guo~\etal propose to encourage prioritization of difficult tasks directly using performance metrics such as accuracy and precision~\cite{dtp}.

\subsection{Knowledge Distillation}

The knowledge distillation aims to transfer the representation capability of a large model (teacher) to a small one (student) to improve its performance~\cite{wei2024scale}. Generally, transferred knowledge can be divided into three types: response-based knowledge, representation-based knowledge, and relation-based knowledge. Response-based knowledge refers to the neural response of the last output layer of the teacher model and it is usually transferred to the student network by matching the soft output distribution of teacher and student networks~\cite{Distilling}. While the idea of response-based knowledge is straightforward and easy to understand, it only considers the output of the last layer and thus fails to address the intermediate-level supervision from the teacher model~\cite{kds}. Therefore some researchers propose the representation-based methods to model the knowledge of intermediate feature maps and transfer it to the student network by minimizing the discrepancy between the value~\cite{hint}, attention map~\cite{hint1,hint2}, and attention project~\cite{atp} of their feature maps. Because the representation-based knowledge only considers specific layers in the teacher model, researchers further introduce relation-based knowledge to model the relationships between different layers or data samples. This knowledge is transferred from the teacher network to the student network by minimizing the discrepancy between the similarity map~\cite{similar}, distribution~\cite{pkt}, inter-data relations~\cite{relation1} of feature pairs from different layers or samples.

\begin{figure}[t]
\centering
\includegraphics[width=1.0\columnwidth]{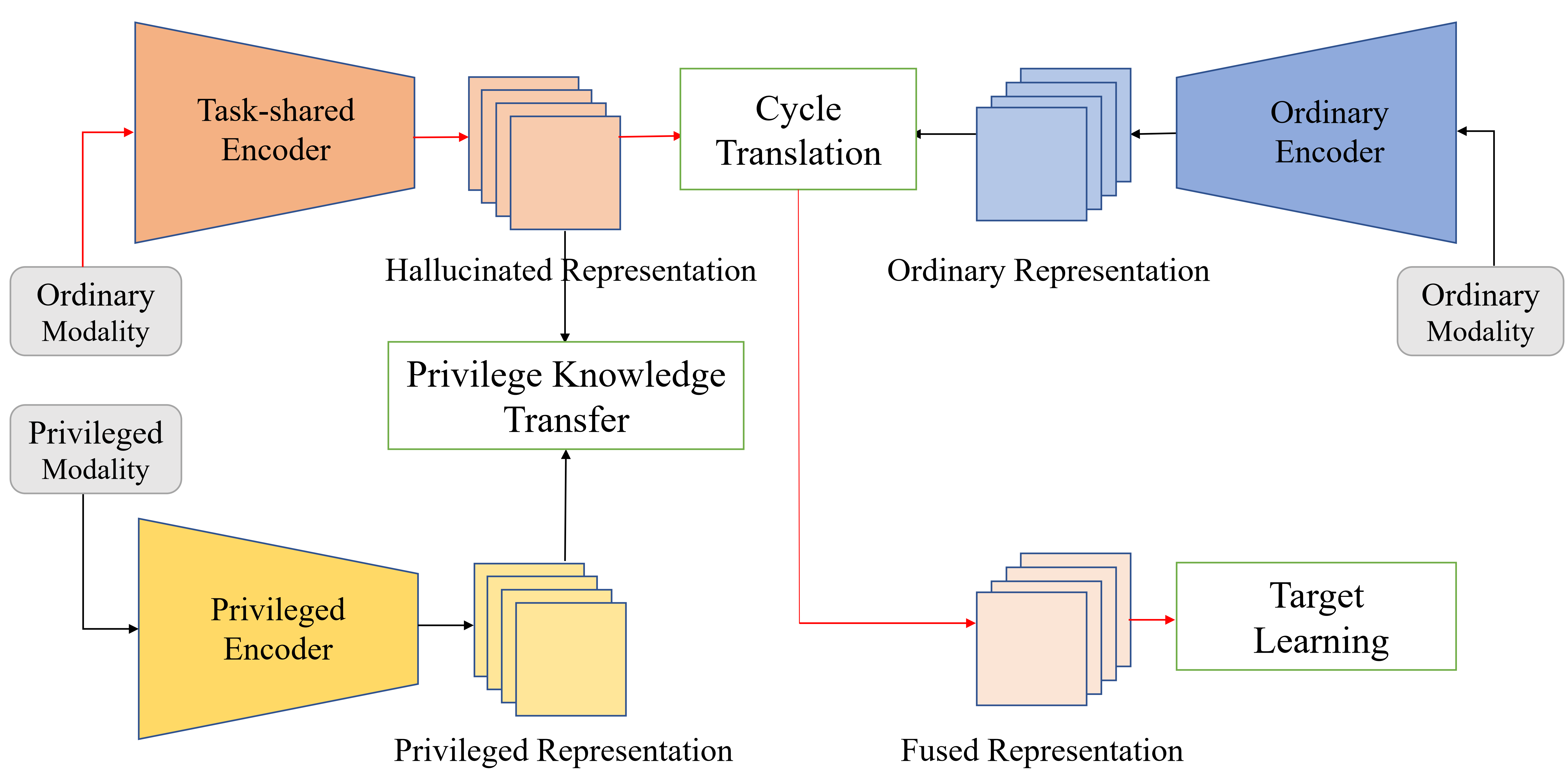} 

\caption{The overall architecture of the one-stage modality distribution framework. It consist of three basic tasks: privileged knowledge transfer, cycle translation, and target learning. The block color means different information flows. The red line means the final inference path.}

\label{framework}
\end{figure}

\section{The one-stage modality-distillation framework}

In this section, we first give an overview of the proposed one-stage modality-distillation (OS-MD) framework that unifies the privileged information transfer and the multiple modalities fusion into a single optimization. Then we introduce the joint adaptation network and cross translation network for the privileged information transfer and the multiple modalities fusion, respectively.

\subsection{Overview}

Existing distillation-based methods perform the hallucination model learning and multiple models fusion independently, which is not conducive for the model to capture the valuable information for the final decision. To tackle this problem, we propose the OS-MD framework to unify these two processes into a single optimization via multitask learning.


Figure~\ref{framework} illustrates the overall architecture of the OS-MD framework. Instead of simply stacking the structures of existing two-stage methods, we summarize three basic tasks and build a general framework via the hard parameter sharing paradigm~\cite{multitask-learning-survey}. Specifically, we draw the hallucination learning procedure as the privileged knowledge transfer task that guides the feature encoder to hallucinate the privilege modality representation from the ordinary modality. This enables the OS-MD framework to utilize more knowledge transfer methods than distillation. Then we decouple the previous modality information fusion procedure into two different tasks: cycle translation and target learning. This enables the OS-MD framework to extract the cross-cues for the final target task. Finally, since these tasks are highly related and the target learning relies on the sample representation refined by the other two tasks, we introduce a task-shared encoder to formula them via the hard parameter sharing paradigm.


Different from conventional multi-task learning that all the tasks are of the same importance, the target learning of the OS-MD framework, such as classification and segmentation, is the primary task while the privileged knowledge transfer and multimodal information tasks are the auxiliary tasks. Therefore, the design of their loss functions is of great significance. Noting that a too difficult auxiliary task would limit the performance of the primary task~\cite{GAN-privilege}. To deal with this challenge, we introduce the joint adaptation network and cross translation network for the privilege knowledge transfer and cycle translation tasks, respectively.

Finally, we introduce the notion for the proposed OS-MD framework here.  let $\mathcal{D}_{o}=\left\{(x_{o}^{(1)},y^{(1)}),...,(x_{o}^{(n)},y^{(n)})\right\}$ and $\mathcal{D}_{p}=\left\{(x_{p}^{(1)},y^{(1)}),...,(x_{p}^{(n)},y^{(n)})\right\}$ denote the datasets for ordinary and privileged modality, respectively. $x_{o}$ and $x_{p}$ are paired mini-batch samples randomly selected from the $\mathcal{D}_{o}$ and $\mathcal{D}_{p}$. ${\varphi_{s}(.)}$, ${\varphi_{p}(.)}$, and ${\varphi_{o}(.)}$ are the transfer function of the task-shared encoder, privileged encoder, and ordinary encoder, respectively. Here, the parameters of the ${\varphi_{o}(.)}$ and ${\varphi_{p}(.)}$ are pretrained by the $\mathcal{D}_{o}$ and $\mathcal{D}_{p}$, respectively. 

\begin{figure}[t]
\centering
\includegraphics[width=1.0\columnwidth]{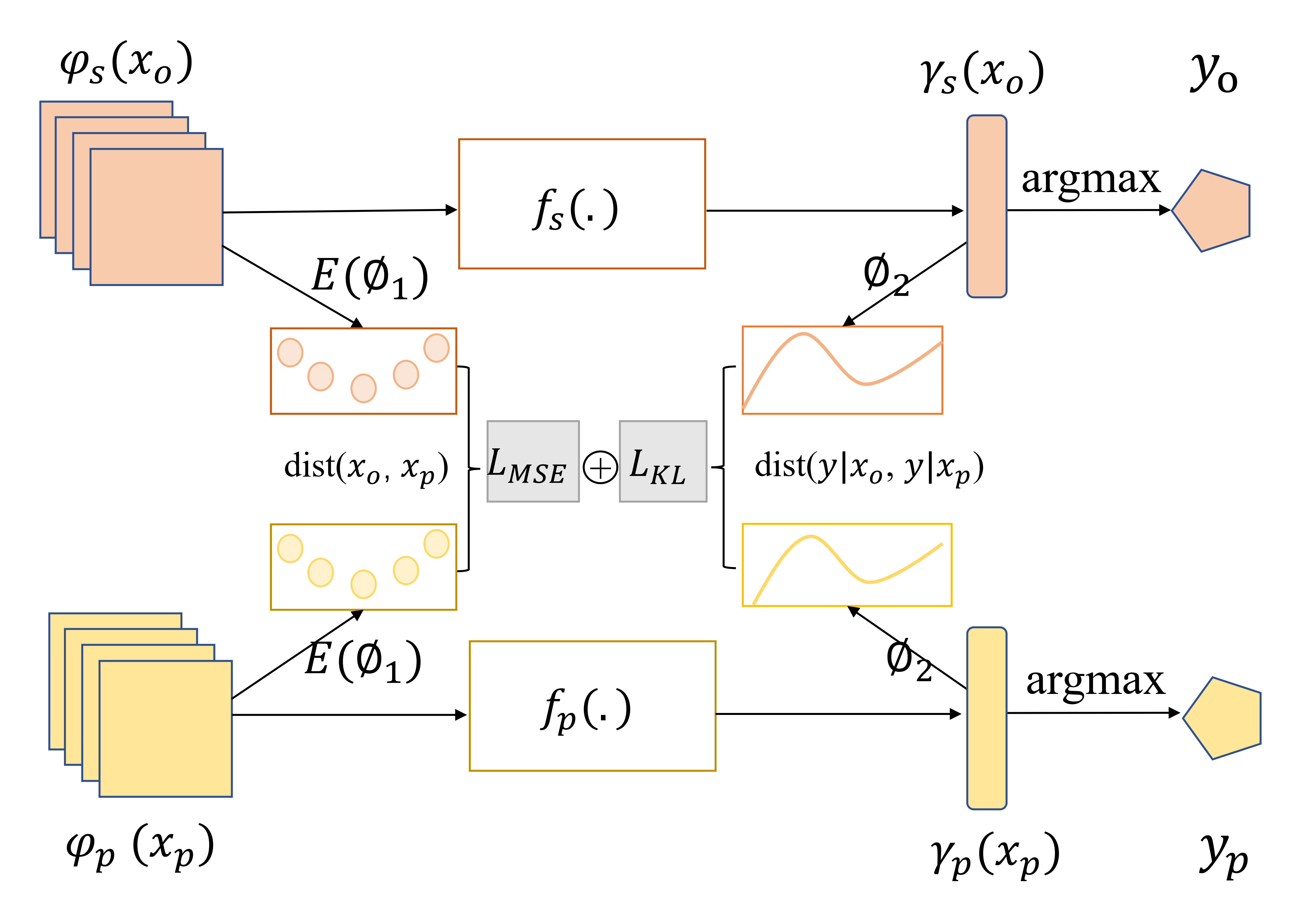} 

\caption{The overall architecture of the joint adaptation network. Its optimization target is to minimize the sum of the distance $dist(x_o,x_p)$ and $dist(y_a|x_o,y_p|x_p)$. The $f_{s}(.)$ and $f_{p}(.)$ are the mapping function from the representation ${\varphi_{s}(x_{o})}$ and ${\varphi_{p}(x_{p})}$ to the logit output $\gamma_{s}(x_{o})$ and $\gamma_{p}(x_{p})$, receptively. $\mathcal{Y}_{o}$ and $\mathcal{Y}_{p}$ are the prediction outputs. $\phi_{1}(.)$ denotes the mapping function that transfers the ${\varphi_{s}(x_{o})}$ and ${\varphi_{p}(x_{p})}$ into reproducing kernel Hilbert space space. $\phi_{2}(.)$ denotes the mapping function that normalizes $\gamma_{s}(x_{o})$ and $\gamma_{p}(x_{p})$ into [0,1]. $E(.)$ denotes the expectation operator. $L_{MSE}$ and $L_{KL}$ denote the MSE and KL loss function, respectively.}

\label{jdnf}
\end{figure}

\subsection{Joint adaptation network}

\label{jdn}

The existing distillation-based methods~\cite{MH1,MH2,ADDA} transfer the privileged knowledge transfer task by matching the intermediate representations of hallucination and privilege models. This may lead to overfitting and limit the performance of the hallucination model since their inputs are different and thus it is not feasible for the hallucination model to generate the same feature map as the privilege model. To tackle this problem, we propose to consider the privileged knowledge transfer task from the perspective of joint distribution adaptation instead of the knowledge distillation and introduce a joint adaptation network (JDN) for this task.



Figure~\ref{jdnf} illustrates the overall architecture of JDN. Here $dist(x_o,x_p)$ means the distance between $Q(\varphi_{s}(x_{o}))$ and $Q(\varphi_{p}(x_{p}))$. $dist(y_a|x_o,y_p|x_p)$ means the distance  between $Q(y|\varphi_{s}(x_{o}))$ and $Q(y|\varphi_{p}(x_{p})$. ${Q(z)}$ denotes the distribution of $z$. Different from the conventional joint distribution adaptation~\cite{jda} that minimizes the sum of the distance between $Q(\varphi_{s}(x_{o}))$ and $Q(\varphi_{s}(x_{p})))$ and the distance between $Q(y|\varphi_{s}(x_{o}))$ and $Q(y|\varphi_{s}(x_{p}))$, JDN further considers the effect of the ${\varphi_{p}(.)}$ and $f_{p}(.)$. This brings more challenges for the optimization.

To calculate the $dist(x_o,x_p)$ and $dist(y_a|x_o,y_p|x_p)$, we need to estimate their distribution function ${Q(z)}$. Although many works have been proposed to solve this problem, accurate estimation is very difficult and complex~\cite{pkt}. This would limit the learning of the primary task. To tackle this problem, we make two innovations. First, inspired by the  maximum mean discrepancy~\cite{mmd}, we propose to approximate the $dist(x_o,x_p)$ with maximum distance of the moment of arbitrary order of $\varphi_{s}(x_{o}$ and $\varphi_{o}(x_{p})$. This avoids the complexity distribution estimation problem and the $dist(x_o,x_p)$ can be rewritten as follows:

\begin{equation}
     dist(x_o,x_p)=\sup_{\|\phi_{1}\|_{\mathcal{H}} \leq 1} |E[\phi_{1}(\varphi_{s}(x_o))]-E[\phi_{1}(\varphi_{p}(x_p))]|^{2}
\end{equation} where, $E[.]$ denote the expectation operator. Furthermore, this can be decomposed via the kernel method~\cite{kernel}:

\begin{equation}
\begin{split}
       dist(x_o,x_p)= &\left\|\frac{1}{b^{2}} \sum_{i}^{b} \sum_{l}^{b} k\left(x_{o}^{i}, x_{o}^{l}\right) -\frac{2}{b^{2}} \sum_{i}^{b} \sum_{j}^{b} k\left(x_{o}^{i}, x_{p}^{j}\right) \right. \\
       &+ \left. \frac{1}{b^{2}} \sum_{j}^{b} \sum_{k}^{b} k\left(x_{p}^{j}, x_{p}^{k}\right)\right\|_{\mathcal{H}}
\end{split}
\end{equation} where $b$ is the batch size of $x_o$ and $x_p$. $k(u,v)$ is the kernel function. Similar with previous study~\cite{mmd}, we set it as the Gaussian function:

\begin{equation}
    k(u, v)=exp({\frac{-\|u-v\|^{2}}{\sigma}})
\end{equation} where exp(.) denotes exponential operation, $\sigma$ denotes the variance.




Second, we propose to calculate the $dist(y_a|x_o,y_p|x_p)$ by the Monte Carlo algorithm. Specifically, the general formula for calculating $dist(y_a|x_o,y_p|x_p)$ is defined as followed:

\begin{equation}
\begin{split}
        dist(y|x_o,y|x_p)=&KL(Q(y|\varphi_{p}(x_{p})||Q(y|\varphi_{s}(x_{o}))) \\
        =&\int Q(y|\varphi_{p}(x_{p}) log\frac{Q(y|\varphi_{p}(x_{p})}{Q(y|\varphi_{s}(x_{o}))}
\end{split}
\end{equation} where $KL(.,.)$ denotes the KL-divergence. According to the Monte Carlo algorithm, we can convert the integral operator to the sampling statistics problems~\cite{pkt} and the $dist(y_a|x_o,y_p|x_p)$ can be rewritten as follows:

\begin{equation}
      dist(y_a|x_o,y_p|x_p)=\sum_{i=1}^{m} \sum_{j=1, i \neq j}^{n} p_{j \mid i} \log \left(\frac{p_{j \mid i}}{q_{j \mid i}}\right)
\end{equation} where $m$ is the number of sample and $n$ is the number of the category corresponding to tasks. $ p_{j \mid i}$ and $q_{j \mid i}$ are the samples drawn from the $Q(y|\varphi_{p}(x_{p})$ and $Q(y|\varphi_{s}(x_{o}))$, respectively. They denotes the conditional probability between $i_{th}$ sample and $j_{th}$ category in the privilege model and hallucination model, respectively.

Thanks to the definition of neural network that the final logits output represents the likelihood that the sample belongs to each category, we can calculate the $ p_{j \mid i}$ and $q_{j \mid i}$ through the $\gamma_{s}(x_{o})$ and $\gamma_{p}(x_{p})$. Furthermore, to satisfy the basic properties of probability, we propose to normalize it using the softmax function $\phi_{2}$. Thus $ p_{j \mid i}$ and $q_{j \mid i}$ can be rewritten as followed:

\begin{equation}
\begin{split}
     &p_{j \mid i}=(\phi_{2}(\gamma_{p}(x_{p})^{i}))_{j} \in[0,1] \\
     &q_{j \mid i}=(\phi_{2}(\gamma_{s}(x_{o})^{i}))_{j} \in [0,1]
\end{split}
\end{equation} where $(.)^{i}$ means the logit output of the $i_th$ sample and $(.)_{j}$ means the $j_{th}$ component of the logit output.

Therefore, the total loss $L_{JDN}$ that guides the training of JDN during a batch can be calculated as follows:
\begin{equation}
    L_{JDN}=  dist(x_o,x_p)+dist(y_a|x_o,y_p|x_p)
\end{equation}

\begin{figure}[t]
\centering
\includegraphics[width=1.0\columnwidth]{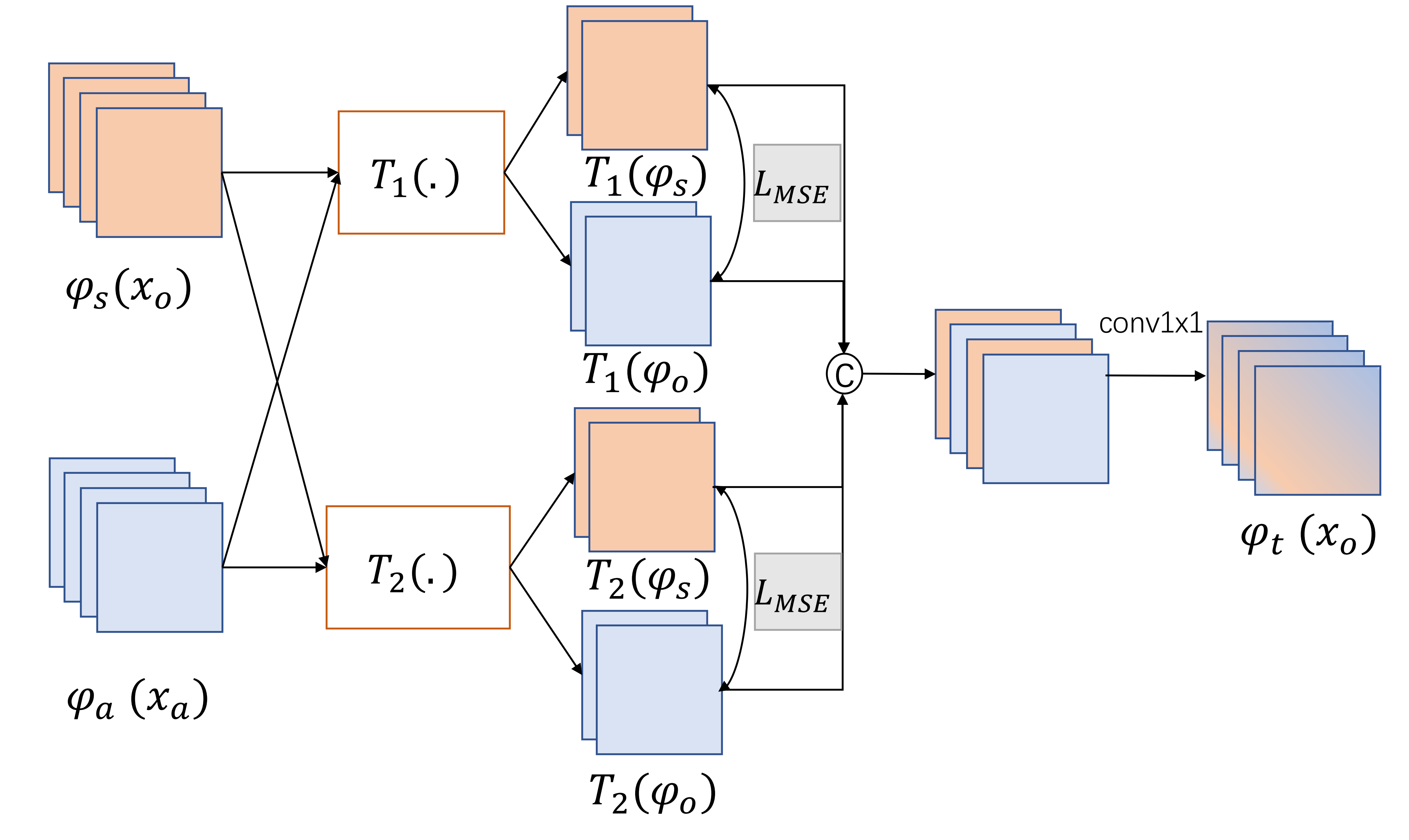} 

\caption{The overall architecture of the cross translation network. $T_{1}(.)$ and $T_{2}(.)$ denote the transform blocks for the hallucinated and ordinary representation $\varphi_{s}(x_{o})$ and $\varphi_{o}(x_{o})$. $\textcircled{c}$ means the concatenation operator.}

\label{cycle}
\end{figure}

\subsection{Cross translation network}
\label{crt}


The existing distillation-based methods~\cite{MH1,MH2,MH3} fuse multimodal information in the output layer by simply adding the logits output of the hallucination and ordinary models. This ignores the intermediate cross-modal cues that are crucial for multimodal learning~\cite{multimodal-survey}. To tackle this problem, we propose the cross translation network (CTN) to translate the hallucinated representation into the ordinary representation. Different from conventional translation methods, CTN performs the representation instead of the sample translation, which reduces the complexity. Furthermore, to extract sufficient cross-modal information, CTN further introduces the parameter sharing strategy to explicitly fuse multimodal information.


Figure~\ref{cycle} illustrates the overview architecture of CTN. $T_{1}(.)$ and $T_{2}(.)$ are implemented with the convolution, batch norm and activation layer. Take $T_{1}(.)$ as an example, it is shared for the input representation $\varphi_{s}(x_{o})$ and target representation $\varphi_{o}(x_{o})$. This not only enables the multiple model information fuse explicitly but also isolates the cross translation network task and the privileged task, so that the learning of the $\varphi_{s}(x_{o})$ can be more flexible. Here, we introduce the reconstruction loss $L_{REC}$ between the transformed feature $T_{1}(\varphi_{s})$ and $T_{1}(\varphi_{o})$, as well as $T_{2}(\varphi_{s})$ and $T_{2}(\varphi_{o})$ via the MSE loss to guide the training of the transform block:

\begin{equation}
    L_{CTN}=L_{MSE}(T_{1}(\varphi_{s}),T_{1}(\varphi_{o}))+L_{MSE}(T_{2}(\varphi_{s}),T_{2}(\varphi_{o}))
\end{equation}

Then we further concatenate all the transformed features and perform a 1x1 convolution to it, so that we can get the final multimodal representation $\varphi_{t}(x_{o})$ for the target task.

\subsection{Training Paradigm}

This section will introduce the crucial details on utilizing the OS-MD framework to address the incomplete multimodal inference, including the total training loss and the balance of multiple losses.

Firstly, the total loss of our OS-MD framework is defined as follows:

\begin{equation}
    L_{total}=\alpha L_{JDN}+\beta L_{CTN}+ \eta L_{TTL}
\end{equation} where $\alpha$, $\beta$, and $\eta$ are the balancing hyper-parameter. $L_{JDN}$ and $L_{CTN}$ have been discussed in the Section~\ref{jdn} and Section~\ref{crt}, respectively. $L_{TTL}$ is the loss that guides the learning of the target task. Because the output of the OS-MD framework is the fusing representation, it can be integrated into arbitrary learning tasks and the $L_{TTL}$ varies with the task. For example, if the target task is the classification, the $L_{TTL}$ can be the cross entropy function.

The balance of multi-task loss has a great influence on the final performance. Although numerous methods have been proposed to tackle this problem, they assume that different tasks are of the same importance. This may not be suitable for the OS-MD framework where the target learning task is the primary task while the privilege knowledge transfer and cycle translation are auxiliary tasks. Therefore, to ensure the performance of the primary task, we set the $\eta$ as 1. Then the weights $\alpha$ and $L_{CTN}$ are dynamically calculated by the gradnorm method~\cite{grad} so that the $L_{JDN}$ and $L_{CTN}$ are maintained at the same magnitude while being an order of magnitude smaller than $L_{TTL}$.

\section{Experiments}

We conduct experiments on the multimodal classification task (face anti-spoofing) and multimodal dense prediction task (segmentation) task to evaluate the performance of the proposed OS-MD framework. In the following, we first compare the proposed OS-MD with the previous state-of-the-arts, ADDA~\cite{ADDA}, MERS~\cite{ADDA}, CMKD~\cite{GAN-privilege}, MT-Net~\cite{MT-Net}, DASK~\cite{DASK}, on the two tasks, respectively. Then, we ablate its important design elements to further study its mechanism. Particularly, all experiments set the general RGB modality as the available modality in the inference like previous work~\cite{MH1,MH2,GAN-privilege,ADDA,MT-Net,gd1}. Still, due to its general design, the proposed framework can also be used to assist other modalities as well.

\begin{table*}[]
\caption{The performance on the RGB-D face anti-spoofing task. The metric is ACER(\%) and the $\downarrow$ means that the lower the value, the better the performance.}
\label{face}
\centering
\begin{tabular}{ccccccc}
\toprule[1pt]
          \multicolumn{1}{c}{Setting}& \multicolumn{1}{c}{Method} & \multicolumn{1}{c}{\begin{tabular}[c]{@{}c@{}}Training\\ Modalities\end{tabular}} & \multicolumn{1}{c}{\begin{tabular}[c]{@{}c@{}}Testing\\ Modalities\end{tabular}} & \multicolumn{1}{c}{CASIA-SURF} ($\downarrow$) & CASIA-CeFA ($\downarrow$) \\ \toprule[1pt]
\multirow{2}{*}{\begin{tabular}[c]{@{}l@{}}Complete\\ Modality\end{tabular}} & SURF            &  RGB                                        & RGB                                        & 10.32      & 36.12     \\ 
          & SURF            &  RGB, D                                     & RGB, D                                     & \textbf{5.53}      & \textbf{30.62}     \\ \hline\hline
\multirow{6}{*}{\begin{tabular}[c]{@{}l@{}}Incomplete\\ Modality\end{tabular}}                                         

          &ADDA           & RGB, D                                      & RGB                                        & 9.83  & 35.50 \\ 
          &MERS           &  RGB, D                                        & RGB                                        & 8.95  & 35.89 \\ 
          &CMKD            &  RGB, D                                      & RGB                                        & 8.64  & 34.20 \\ 
          
         & MT-Net         &  RGB, D                                  & RGB                                        & 8.45      & 33.60     \\

          & DASK         &  RGB, D                                   & RGB                                        & 8.08      & 32.64     \\

          & OS-MD \textbf{(Ours)}            &  RGB, D                                    & RGB                                        & \textbf{6.75}      & \textbf{31.26}   \\       
          \toprule[1pt]
          
\end{tabular}

\end{table*}

\subsection{Multimodal classification on the face anti-spoofing task}
\label{face-anti}
\textbf{Settings}: For multimodal face anti-spoofing, we report the result on the CASIA-SURF~\cite{surf} and CASIA-CeFA~\cite{cefa} datasets with the intra-testing protocol as well as cross-ethnicity and cross-attack protocol suggested by authors, respectively. For a fair comparison, we follow the same benchmark model and data augmentation strategy with CASIA-SURF~\cite{surf}. The models are trained with the SGD optimizer, where the learning rate is 0.001, momentum is 0.9, batch size is 64, and maximum epoch is 50. The metric used is Average Classification Error Rate (ACER)~\cite{surf}.

\textbf{Result and analysis}: Table~\ref{face} shows the experiment results on the multimodal face anti-spoofing task. From this table, we have the following observations:

(1) While the MT-Net~\cite{MT-Net} trains the hallucination model by the complex imputation-based methods, it unifies the sample reconstruction and multimodal fusion into a single optimizer and outperforms the CMKD~\cite{GAN-privilege}, MERS~\cite{ADDA}, and ADDA~\cite{ADDA} methods that train the hallucination model by relation-based, feature-based, and response-based algorithms, respectively. This demonstrates the potential of one-stage learning.

(2) The DASK~\cite{DASK} is recently proposed to train the hallucination model by transferring the global and local knowledge and outperforms the MT-Net~\cite{MT-Net}. This again demonstrates the effectiveness of distillation-based methods that transfer the privileged knowledge in the low-dimensional feature space.

(3) The proposed OS-MD method further outperforms the DASK and achieves state-of-the-art performance. For example, it exceeds DASK by 1.33\% and 1.38\% in the CASIA-SURF and CASIA-CeFA dataset, respectively. This verifies the superiority to joint hallucination learning and multimodal information into a single optimization procedure.

Overall, the proposed OS-MD method outperforms the baselines, and the literature advances consistently, verifying its superiority on the multimodal classification task.



\begin{table*}[]
\centering
\caption{The performance on the RGB-D semantic segmentation task. The metric is MIOU(\%) and the $\uparrow$ means that the higher the value, the better the performance. The default backbone is ResNet34 and the suffix `-R50' means that the backbone is ResNet50.}
\label{seg}
\begin{tabular}{cccccc}
\hline
Setting                                                                        & Method & \begin{tabular}[c]{@{}c@{}}Training\\ Modalities\end{tabular} & \begin{tabular}[c]{@{}c@{}}Testing\\ Modalities\end{tabular} & NYUV2 ($\uparrow$) & SUN RGBD ($\uparrow$) \\ \toprule[1pt]
\multirow{2}{*}{\begin{tabular}[c]{@{}c@{}}Complte\\ Modality\end{tabular}}    & ESANet & RGB                                                           & RGB                                                          & 42.96 & 43.46    \\ 
                                                     & ESANet & RGB, D                                                          & RGB, D                                                         & 48.97 & 48.87    \\ 
            & ESANet-R50 & RGB                                                          & RGB                                                         & 44.22 & 42.85    \\                                                                   \toprule[1pt]
\multirow{4}{*}{\begin{tabular}[c]{@{}c@{}}Incomplete\\ Modality\end{tabular}} & ADDA   & RGB, D                                                          & RGB                                                          & 43.57 & 41.81    \\ 
                                                                               & CMKD   & RGB, D                                                          & RGB                                                          & 44.03 & 42.35    \\ 
                                                                               & DASK  & RGB, D                                                          & RGB                                                          & 44.45 & 42.66    \\
                                                                               & OS-MD  & RGB, D                                                          & RGB                                                          & \textbf{45.55} & \textbf{43.64}    \\ \toprule[1pt]
\end{tabular}

\end{table*}

\subsection{Multimodal dense prediction on the Semantic segmentation task}

\textbf{Settings}: For the multimodal semantic segmentation task, we report the result on the NYUv2~\cite{nyuv2} and the SUN RGBD~\cite{sunrgb} datasets. Here we use the common 40-class label setting of the NYUv2. The baseline method is set as the latest efficient scene analysis Network (ESANet)~\cite{ESANet}. For fair comparisons, we follow the same benchmark model and data augmentation strategy with the ESANet~\cite{ESANet}. Specifically, we choose the ResNet34 and ResNet50 as the backbone network, respectively. The models are trained with the SGD optimizer, where the initial learning rate is 0.01, momentum is 0.9, batch size is 8, and maximum epoch is 300. Besides, the one-cycle learning rate scheduler is used to adapt the learning rate. The metric used is mean intersection over union (MIOU)~\cite{ESANet}.

\textbf{Results and Analysis}: Table~\ref{seg} shows the experiment results on the multimodal semantic segmentation task. From this table, we have the following observations:

(1)The OS-MD method outperforms both the basic ESANet and the DASK by 2.36\% and 1.10\% in NYUV2; by 2.58\% and 0.94\% in SUN RGBD, respectively. This demonstrates its effectiveness in the dense prediction task.

(2)When only the RGB modality is available for inference, the OS-MD implemented with ResNet34 even outperforms the conventional ESANet implemented with ResNet50 by 1.01\% in NYUv2 and 0.79\% in SUN RGBD. This shows the potential of the OS-MD to achieve better performance while maintaining low complexity.

Overall, the proposed OS-MD method outperforms the baselines and the latest privilege learning method, validating its scalability to dense prediction tasks.

\subsection{Ablation Study}

In this section, we ablate important design elements in the proposed method. To study both the JDN and CRT strategies, we report the result on the CASIA-SURF dataset. The experiment settings are the same as those in Section~\ref{face-anti}.

\begin{table}[]
\caption{The performance on the RGB-IR face anti-spoofing task. The metric is ACER(\%) and the $\downarrow$ means that the lower the value, the better the performance.}
\label{rgbir}
\begin{tabular}{cccc}
 \toprule[1pt]
\multicolumn{1}{l}{Method} & \begin{tabular}[c]{@{}c@{}}Traing\\ Modalities\end{tabular} & \begin{tabular}[c]{@{}c@{}}Testing\\ Modalities\end{tabular} & ACER($\downarrow$)            \\  \toprule[1pt]
\multirow{1}{*}{SURF}        & \multirow{1}{*}{RGB}                                     & \multirow{1}{*}{RGB}                                         & \multirow{1}{*}{10.32}       \\ 
\multirow{1}{*}{SURF}        & \multirow{1}{*}{RGB,IR}                                     & \multirow{1}{*}{RGB,IR}                                         & \multirow{1}{*}{6.32}       \\   \toprule[1pt]
DASK                        & RGB,IR                                                      & RGB                                                          &          8.52         \\ 
\textbf{OS-MD(ours)}         & RGB,IR                                                      & RGB                                                          &       7.64            \\  \toprule[1pt]
\end{tabular}

\end{table}

\subsubsection{The scalability to other modal composition} To study the effectiveness of the OD-MD method in other modality compositions, such as RGB and IR modalities. We perform the experiment on the CASIA-SURF dataset and assume that its IR modality is only available during the training stage. As shown in the Table~\ref{rgbir}, the OS-MD method outperforms the SURF and DASK methods by 2.68\% and 0.88\%, respectively, when the IR modality is missing during the inference stage.

\begin{table}[]
\caption{The ablation experiments for the JDN strategy. We report the result of the original OS-MD framework (Full), the OS-MD framework without the JDN (-JDN), the OS-MD framework with the feature-based distillation algorithm (-JDN+FKD), and the OS-MD framework with the relation-based distillation algorithm (-JDN+RKD).}

\label{JDN}
\centering
\begin{tabular}{cccc}
\toprule[1pt]
Full             \quad          & -JDN \quad& -JDN+FKD        \quad          & -JDN+RKD       \quad            \\ \hline
\multicolumn{1}{c}{6.75}\quad & 44.76\quad & \multicolumn{1}{c}{8.14}\quad & 7.65  \\ \toprule[1pt]
\end{tabular}

\end{table}

\subsubsection{Study of the joint adaptation network} Here we ablate the joint adaptation network module of the OS-MD framework to evaluate their impacts.  Table~\ref{JDN} shows the results. The OS-MD framework without joint adaptation network degrades the performance of the complete one by 38.01\% in the CASIA-SURF. This demonstrates the importance of privileged knowledge transfer to the OS-MD framework. The OS-MD framework with the feature-based distilltion and relation-based distillation algorithm degrades the performance of the complete one by 1.39\% and 0.9\%, respoectivly. This demonstrate the effectiveness to transfer the privilege knowledge by matching the marginal and conditional probability distributions simultaneously.

\begin{table}[]
\caption{The ablation experiments for the CTN strategy. We report the result of the original OS-MD framework (Full), the OS-MD framework without CTN (-CTN), and the OS-MD framework with auto-encoder (-CTN+AE).}

\label{CTN}
\centering
\begin{tabular}{ccc}
\toprule[1pt]
Full            \quad      & -CTN \quad& -CTN+AE        \quad     \\ \hline
\multicolumn{1}{c}{6.75}\quad & 9.16\quad & \multicolumn{1}{c}{8.67}  \\ \toprule[1pt]
\end{tabular}

\end{table}

\subsubsection{Study of the cross translation network} Here we ablate the cross translation network module of the OS-MD framework to evaluate their impacts.  Table~\ref{CTN} shows the results. The OS-MD framework without the cross translation network degrades the performance of the complete one by 2.41\% in the CASIA-SURF. This demonstrates the importance of cross translation network. The OS-MD framework with the auto-encoder algorithm degrades the performance of the complete one by 1.92\%. This demonstrates the effectiveness to perform the representation translation via the sharing parameters.

\section{Conclusion}

In this paper, we propose an OS-MD framework to assist the incomplete multimodal inference. It performs the hallucination learning and multiple model fusion simultaneously so that they can negotiate with each other to capture valuable privileged knowledge. Instead of simply stacking the structures of existing two-stage methods, we draw three basic tasks: privilege knowledge transfer, cycle translation, and target learning, and organize them via hard parameter sharing. Then we propose two strategies called JDN and CTN for the privilege knowledge transfer and cycle translation tasks, respectively. The JDN guides the hallucination model to learn from the privilege model by matching their marginal and conditional probability distributions simultaneously. This is capable of overcoming representational heterogeneity caused by the input discrepancy while preserving the valuable decision boundary. The CTN proposes to translate the hallucinated representation into the ordinary representation by sharing parameters. This help to fuse the multimodal information explicitly. Finally, extensive experiments on RGB-D classification and segmentation tasks well demonstrate the superiority and generalization of our algorithms.

{\small
\bibliographystyle{ieee_fullname}
\bibliography{egbib}

\begin{thebibliography}{10}\itemsep=-1pt

\bibitem{multimodal-survey}
Tadas Baltru{\v{s}}aitis, Chaitanya Ahuja, and Louis-Philippe Morency.
\newblock Multimodal machine learning: A survey and taxonomy.
\newblock {\em TPAMI}, 41(2):423--443, 2018.

\bibitem{d2d}
Jonathan C, David~A Stroud, Chen Ross, Jia Sun, Rahul Deng, and Sukthankar.
\newblock D3d: Distilled 3d networks for video action recognition.
\newblock In {\em Proceedings of the IEEE/CVF Winter Conference on Applications
  of Computer Vision}, pages 2755--2764, 2020.

\bibitem{gan-issue1}
Lei Cai, Zhengyang Wang, Hongyang Gao, Dinggang Shen, and Shuiwang Ji.
\newblock Deep adversarial learning for multi-modality missing data completion.
\newblock In {\em Proceedings of the ACM SIGKDD International Conference on
  Knowledge Discovery \& Data Mining}, pages 1158--1166, 2018.

\bibitem{rgbd-seg-1}
Jinming Cao, Hanchao Leng, Dani Lischinski, Daniel Cohen-Or, Changhe Tu, and
  Yangyan Li.
\newblock Shapeconv: Shape-aware convolutional layer for indoor rgb-d semantic
  segmentation.
\newblock In {\em Proceedings of the IEEE/CVF International Conference on
  Computer Vision}, pages 7088--7097, 2021.

\bibitem{grad}
Zhao Chen, Vijay Badrinarayanan, Chen-Yu Lee, and Andrew Rabinovich.
\newblock Gradnorm: Gradient normalization for adaptive loss balancing in deep
  multitask networks.
\newblock In {\em International Conference on Machine Learning}, pages
  794--803. PMLR, 2018.

\bibitem{MH2}
Nuno~C Garcia, Pietro Morerio, and Vittorio Murino.
\newblock Modality distillation with multiple stream networks for action
  recognition.
\newblock In {\em Proceedings of the European Conference on Computer Vision},
  pages 103--118, 2018.

\bibitem{GAN-privilege}
Nuno~C Garcia, Pietro Morerio, and Vittorio Murino.
\newblock Learning with privileged information via adversarial discriminative
  modality distillation.
\newblock {\em IEEE transactions on pattern analysis and machine intelligence},
  42(10):2581--2593, 2019.

\bibitem{ADDA}
Nuno~C Garcia, Pietro Morerio, and Vittorio Murino.
\newblock Learning with privileged information via adversarial discriminative
  modality distillation.
\newblock {\em IEEE transactions on pattern analysis and machine intelligence},
  42(10):2581--2593, 2019.

\bibitem{kds}
Jianping Gou, Baosheng Yu, Stephen~J Maybank, and Dacheng Tao.
\newblock Knowledge distillation: A survey.
\newblock {\em International Journal of Computer Vision}, 129(6):1789--1819,
  2021.

\bibitem{dtp}
Michelle Guo, Albert Haque, De-An Huang, Serena Yeung, and Li Fei-Fei.
\newblock Dynamic task prioritization for multitask learning.
\newblock In {\em Proceedings of the European conference on computer vision
  (ECCV)}, pages 270--287, 2018.

\bibitem{Distilling}
Geoffrey Hinton, Oriol Vinyals, and Jeff Dean.
\newblock Distilling the knowledge in a neural network.
\newblock {\em arXiv preprint arXiv:1503.02531}, 2015.

\bibitem{MH1}
Judy Hoffman, Saurabh Gupta, and Trevor Darrell.
\newblock Learning with side information through modality hallucination.
\newblock In {\em Proceedings of the IEEE Conference on Computer Vision and
  Pattern Recognition}, pages 826--834, 2016.

\bibitem{discri3}
Danfeng Hong, Jocelyn Chanussot, Naoto Yokoya, Jian Kang, and Xiao~Xiang Zhu.
\newblock Learning-shared cross-modality representation using
  multispectral-lidar and hyperspectral data.
\newblock {\em IEEE Geoscience and Remote Sensing Letters}, 17(8):1470--1474,
  2020.

\bibitem{hint2}
Zehao Huang and Naiyan Wang.
\newblock Like what you like: Knowledge distill via neuron selectivity
  transfer.
\newblock {\em arXiv preprint arXiv:1707.01219}, 2017.

\bibitem{privacy1}
Mimansa Jaiswal and Emily~Mower Provost.
\newblock Privacy enhanced multimodal neural representations for emotion
  recognition.
\newblock In {\em Proceedings of the AAAI Conference on Artificial
  Intelligence}, volume~34, pages 7985--7993, 2020.

\bibitem{action_multi_2}
Yu-Gang Jiang, Zuxuan Wu, Jinhui Tang, Zechao Li, Xiangyang Xue, and Shih-Fu
  Chang.
\newblock Modeling multimodal clues in a hybrid deep learning framework for
  video classification.
\newblock {\em IEEE Transactions on Multimedia}, 20(11):3137--3147, 2018.

\bibitem{MC2}
Jiang Jue, Hu Jason, Tyagi Neelam, Rimner Andreas, Berry~L Sean, Deasy~O
  Joseph, and Veeraraghavan Harini.
\newblock Integrating cross-modality hallucinated mri with ct to aid
  mediastinal lung tumor segmentation.
\newblock In {\em International Conference on Medical Image Computing and
  Computer-Assisted Intervention}, pages 221--229. Springer, 2019.

\bibitem{pixel-based-3}
Jiang Jue, Hu Jason, Tyagi Neelam, Rimner Andreas, Berry~L Sean, Deasy~O
  Joseph, and Veeraraghavan Harini.
\newblock Integrating cross-modality hallucinated mri with ct to aid
  mediastinal lung tumor segmentation.
\newblock In {\em International Conference on Medical Image Computing and
  Computer-Assisted Intervention}, pages 221--229. Springer, 2019.

\bibitem{hint1}
Nikos Komodakis and Sergey Zagoruyko.
\newblock Paying more attention to attention: improving the performance of
  convolutional neural networks via attention transfer.
\newblock In {\em International Conference on Learning Representations}, 2017.

\bibitem{miss-hall2}
Saurabh Kumar, Biplab Banerjee, and Subhasis Chaudhuri.
\newblock Improved landcover classification using online spectral data
  hallucination.
\newblock {\em Neurocomputing}, 439:316--326, 2021.

\bibitem{relation1}
Seunghyun Lee and Byung~Cheol Song.
\newblock Graph-based knowledge distillation by multi-head attention network.
\newblock {\em arXiv preprint arXiv:1907.02226}, 2019.

\bibitem{miss-dhad}
Xiao Li, Lin Lei, Yuli Sun, and Gangyao Kuang.
\newblock Dynamic-hierarchical attention distillation with synergetic instance
  selection for land cover classification using missing heterogeneity images.
\newblock {\em IEEE Transactions on Geoscience and Remote Sensing}, 60:1--16,
  2021.

\bibitem{MH3}
Xiao Li, Lin Lei, Yuli Sun, and Gangyao Kuang.
\newblock Dynamic-hierarchical attention distillation with synergetic instance
  selection for land cover classification using missing heterogeneity images.
\newblock {\em IEEE Transactions on Geoscience and Remote Sensing}, 60:1--16,
  2021.

\bibitem{cefa}
Ajian Liu, Zichang Tan, Jun Wan, Sergio Escalera, Guodong Guo, and Stan~Z Li.
\newblock Casia-surf cefa: A benchmark for multi-modal cross-ethnicity face
  anti-spoofing.
\newblock In {\em Proceedings of the IEEE/CVF Winter Conference on Applications
  of Computer Vision}, pages 1179--1187, 2021.

\bibitem{MT-Net}
Ajian Liu, Zichang Tan, Jun Wan, Yanyan Liang, Zhen Lei, Guodong Guo, and
  Stan~Z Li.
\newblock Face anti-spoofing via adversarial cross-modality translation.
\newblock {\em IEEE Transactions on Information Forensics and Security},
  16:2759--2772, 2021.

\bibitem{gan-issue2}
Haojie Liu, Shun Ma, Daoxun Xia, and Shaozi Li.
\newblock Sfanet: A spectrum-aware feature augmentation network for
  visible-infrared person re-identification.
\newblock {\em arXiv preprint arXiv:2102.12137}, 2021.

\bibitem{action_multi_all_1}
Mengyuan Liu and Junsong Yuan.
\newblock Recognizing human actions as the evolution of pose estimation maps.
\newblock In {\em Proceedings of the IEEE Conference on Computer Vision and
  Pattern Recognition}, pages 1159--1168, 2018.

\bibitem{jda}
Mingsheng Long, Jianmin Wang, Guiguang Ding, Jiaguang Sun, and Philip~S Yu.
\newblock Transfer feature learning with joint distribution adaptation.
\newblock In {\em Proceedings of the IEEE international conference on computer
  vision}, pages 2200--2207, 2013.

\bibitem{gd1}
Zelun Luo, Jun-Ting Hsieh, Lu Jiang, Juan~Carlos Niebles, and Li Fei-Fei.
\newblock Graph distillation for action detection with privileged modalities.
\newblock In {\em Proceedings of the European Conference on Computer Vision
  (ECCV)}, pages 166--183, 2018.

\bibitem{kernel}
Krikamol Muandet, Kenji Fukumizu, Bharath Sriperumbudur, Bernhard
  Sch{\"o}lkopf, et~al.
\newblock Kernel mean embedding of distributions: A review and beyond.
\newblock {\em Foundations and Trends{\textregistered} in Machine Learning},
  10(1-2):1--141, 2017.

\bibitem{DASK}
Jianyuan Ni, Raunak Sarbajna, Yang Liu, Anne~HH Ngu, and Yan Yan.
\newblock Cross-modal knowledge distillation for vision-to-sensor action
  recognition.
\newblock In {\em ICASSP 2022-2022 IEEE International Conference on Acoustics,
  Speech and Signal Processing (ICASSP)}, pages 4448--4452. IEEE, 2022.

\bibitem{mmd}
Sinno~Jialin Pan, Ivor~W Tsang, James~T Kwok, and Qiang Yang.
\newblock Domain adaptation via transfer component analysis.
\newblock {\em IEEE transactions on neural networks}, 22(2):199--210, 2010.

\bibitem{MC1}
Yongsheng Pan, Mingxia Liu, Chunfeng Lian, Yong Xia, and Dinggang Shen.
\newblock Spatially-constrained fisher representation for brain disease
  identification with incomplete multi-modal neuroimages.
\newblock {\em IEEE Transactions on Medical Imaging}, 39(9):2965--2975, 2020.

\bibitem{pixel-based-2}
Yongsheng Pan, Mingxia Liu, Chunfeng Lian, Yong Xia, and Dinggang Shen.
\newblock Spatially-constrained fisher representation for brain disease
  identification with incomplete multi-modal neuroimages.
\newblock {\em IEEE Transactions on Medical Imaging}, 39(9):2965--2975, 2020.

\bibitem{pkt}
Nikolaos Passalis and Anastasios Tefas.
\newblock Learning deep representations with probabilistic knowledge transfer.
\newblock In {\em Proceedings of the European Conference on Computer Vision},
  pages 268--284, 2018.

\bibitem{atp}
Peyman Passban, Yimeng Wu, Mehdi Rezagholizadeh, and Qun Liu.
\newblock Alp-kd: Attention-based layer projection for knowledge distillation.
\newblock {\em arXiv preprint arXiv:2012.14022}, 2020.

\bibitem{rgbd-seg-3}
Yeqiang Qian, Liuyuan Deng, Tianyi Li, Chunxiang Wang, and Ming Yang.
\newblock Gated-residual block for semantic segmentation using rgb-d data.
\newblock {\em IEEE Transactions on Intelligent Transportation Systems}, 2021.

\bibitem{hint}
Adriana Romero, Nicolas Ballas, Samira~Ebrahimi Kahou, Antoine Chassang, Carlo
  Gatta, and Yoshua Bengio.
\newblock Fitnets: Hints for thin deep nets.
\newblock {\em arXiv preprint arXiv:1412.6550}, 2014.

\bibitem{rgbd-seg-2}
Daniel Seichter, Mona K{\"o}hler, Benjamin Lewandowski, Tim Wengefeld, and
  Horst-Michael Gross.
\newblock Efficient rgb-d semantic segmentation for indoor scene analysis.
\newblock In {\em 2021 IEEE International Conference on Robotics and Automation
  (ICRA)}, pages 13525--13531. IEEE, 2021.

\bibitem{ESANet}
Daniel Seichter, Mona K{\"o}hler, Benjamin Lewandowski, Tim Wengefeld, and
  Horst-Michael Gross.
\newblock Efficient rgb-d semantic segmentation for indoor scene analysis.
\newblock In {\em 2021 IEEE International Conference on Robotics and Automation
  (ICRA)}, pages 13525--13531. IEEE, 2021.

\bibitem{a-attack2}
Ali Shafahi, Mahyar Najibi, Zheng Xu, John Dickerson, Larry~S Davis, and Tom
  Goldstein.
\newblock Universal adversarial training.
\newblock In {\em Proceedings of the AAAI Conference on Artificial
  Intelligence}, volume~34, pages 5636--5643, 2020.

\bibitem{nyuv2}
Nathan Silberman, Derek Hoiem, Pushmeet Kohli, and Rob Fergus.
\newblock Indoor segmentation and support inference from rgbd images.
\newblock In {\em European conference on computer vision}, pages 746--760.
  Springer, 2012.

\bibitem{sunrgb}
Shuran Song, Samuel~P Lichtenberg, and Jianxiong Xiao.
\newblock Sun rgb-d: A rgb-d scene understanding benchmark suite.
\newblock In {\em Proceedings of the IEEE conference on computer vision and
  pattern recognition}, pages 567--576, 2015.

\bibitem{similar}
Frederick Tung and Greg Mori.
\newblock Similarity-preserving knowledge distillation.
\newblock In {\em Proceedings of the IEEE/CVF International Conference on
  Computer Vision}, pages 1365--1374, 2019.

\bibitem{multitask-learning-survey}
Simon Vandenhende, Stamatios Georgoulis, Wouter Van~Gansbeke, Marc Proesmans,
  Dengxin Dai, and Luc Van~Gool.
\newblock Multi-task learning for dense prediction tasks: A survey.
\newblock {\em IEEE transactions on pattern analysis and machine intelligence},
  2021.

\bibitem{mmanet}
Shicai Wei, Chunbo Luo, and Yang Luo.
\newblock Mmanet: Margin-aware distillation and modality-aware regularization
  for incomplete multimodal learning.
\newblock In {\em Proceedings of the IEEE/CVF Conference on Computer Vision and
  Pattern Recognition}, pages 20039--20049, 2023.

\bibitem{wei2024scale}
Shicai Wei, Chunbo Luo, and Yang Luo.
\newblock Scale decoupled distillation.
\newblock In {\em Proceedings of the IEEE/CVF Conference on Computer Vision and
  Pattern Recognition}, pages 15975--15983, 2024.

\bibitem{wei2025boosting}
Shicai Wei, Chunbo Luo, and Yang Luo.
\newblock Boosting multimodal learning via disentangled gradient learning.
\newblock {\em arXiv preprint arXiv:2507.10213}, 2025.

\bibitem{wei2025improving}
Shicai Wei, Chunbo Luo, and Yang Luo.
\newblock Improving multimodal learning via imbalanced learning.
\newblock {\em arXiv preprint arXiv:2507.10203}, 2025.

\bibitem{wei2023privileged}
Shicai Wei, Chunbo Luo, Yang Luo, and Jialang Xu.
\newblock Privileged modality learning via multimodal hallucination.
\newblock {\em IEEE Transactions on Multimedia}, 2023.

\bibitem{wei2024gradient}
Shicai Wei, Chunbo Luo, Xiaoguang Ma, and Yang Luo.
\newblock Gradient decoupled learning with unimodal regularization for
  multimodal remote sensing classification.
\newblock {\em IEEE Transactions on Geoscience and Remote Sensing}, 2024.

\bibitem{wei2023msh}
Shicai Wei, Yang Luo, Xiaoguang Ma, Peng Ren, and Chunbo Luo.
\newblock Msh-net: Modality-shared hallucination with joint adaptation
  distillation for remote sensing image classification using missing
  modalities.
\newblock {\em IEEE Transactions on Geoscience and Remote Sensing}, 61:1--15,
  2023.

\bibitem{dmr}
Shicai Wei, Yang Luo, Yuji Wang, and Chunbo Luo.
\newblock Robust multimodal learning via representation decoupling.
\newblock In {\em European Conference on Computer Vision}, pages 38--54.
  Springer, 2025.

\bibitem{a-attack1}
Valentina Zantedeschi, Maria-Irina Nicolae, and Ambrish Rawat.
\newblock Efficient defenses against adversarial attacks.
\newblock In {\em Proceedings of the 10th ACM Workshop on Artificial
  Intelligence and Security}, pages 39--49, 2017.

\bibitem{surf}
Shifeng Zhang, Xiaobo Wang, Ajian Liu, Chenxu Zhao, Jun Wan, Sergio Escalera,
  Hailin Shi, Zezheng Wang, and Stan~Z Li.
\newblock A dataset and benchmark for large-scale multi-modal face
  anti-spoofing.
\newblock In {\em Proceedings of the IEEE/CVF Conference on Computer Vision and
  Pattern Recognition}, pages 919--928, 2019.

\bibitem{cyclegan}
Jun-Yan Zhu, Taesung Park, Phillip Isola, and Alexei~A Efros.
\newblock Unpaired image-to-image translation using cycle-consistent
  adversarial networks.
\newblock In {\em Proceedings of the IEEE international conference on computer
  vision}, pages 2223--2232, 2017.

\end{thebibliography}
}





\end{document}